\documentclass[sigconf,screen]{acmart}

\def\BibTeX{{\rm B\kern-.05em{\sc i\kern-.025em b}\kern-.08em
    T\kern-.1667em\lower.7ex\hbox{E}\kern-.125emX}}

\AtBeginDocument{%
  \providecommand\BibTeX{{%
    \normalfont B\kern-0.5em{\scshape i\kern-0.25em b}\kern-0.8em\TeX}}}

\copyrightyear{2025}
\acmYear{2025}
\setcopyright{cc}
\setcctype{by}
\acmConference[FSE Companion '25]{33rd ACM International Conference on the Foundations of Software Engineering}{June 23--28, 2025}{Trondheim, Norway}
\acmBooktitle{33rd ACM International Conference on the Foundations of Software Engineering (FSE Companion '25), June 23--28, 2025, Trondheim, Norway}\acmDOI{10.1145/3696630.3728572}
\acmISBN{979-8-4007-1276-0/2025/06}

\usepackage{multirow}
\usepackage{subcaption}
\usepackage{tabularx}
\usepackage{soul}
\usepackage{xcolor}
\usepackage{tcolorbox}
\usepackage{wrapfig}
\usepackage{pifont}
\usepackage{makecell}
\usepackage{orcidlink}
\usepackage[shortlabels]{enumitem}

\newcommand\innerwidth{2mm}

\newcommand{\cmark}{\ding{51}}%
\newcommand{\xmark}{\ding{55}}%
\newcommand{\ncmark}{\ding{51}\kern-1.1ex\raisebox{.7ex}{\rotatebox[origin=c]{125}{--}}}

\newcommand{\modelcount}{24 }

\newcommand{\mccount}{7,334 }
\newcommand{\qacount}{1,736 }
\newcommand{\allquestioncount}{9,070 }

\newcommand{\companycount}{11 }

\newcommand{\biased}[1]{
  \begingroup
  \sethlcolor{red}
  {\hl{#1}}
  \endgroup
}

\newcommand{\unbiased}[1]{
  \begingroup
  \sethlcolor{green}
  {\hl{#1}}
  \endgroup
}

\newenvironment{insight}{
    \begin{center}
    \begin{tcolorbox}[colback=blue!5,
      colframe=gray!10,
      arc=2mm, auto outer arc,
      boxrule=0.5pt,
      left=\innerwidth,
      right=\innerwidth,
    ]
}
{
    \end{tcolorbox}
    \end{center}
}

\begin{document}

\title{OpsEval: A Comprehensive Benchmark Suite for Evaluating Large Language Models’ Capability in IT Operations Domain}
\renewcommand{\shorttitle}{OpsEval: A Comprehensive IT Operations Benchmark Suite}


\settopmatter{authorsperrow=5}

\author{Yuhe Liu} \orcid{0009-0002-0324-7749}
\affiliation{
    \institution{Tsinghua University \& BNRist}
    \city{Beijing} \country{China}
}

\author{Changhua Pei} \orcid{0000-0001-9288-4787}
\affiliation{
    \institution{CNIC, CAS}
    \city{Beijing} \country{China}
}

\author{Longlong Xu} \orcid{0009-0008-7364-1307}
\author{Bohan Chen} \orcid{0009-0002-6804-7473}
\affiliation{
    \institution{Tsinghua University \& BNRist}
    \city{Beijing} \country{China}
}

\author{Mingze Sun} \orcid{0000-0002-4205-7182}
\affiliation{
    \institution{Tsinghua University \& BNRist}
    \city{Beijing} \country{China}
}

\author{Zhirui Zhang} \orcid{0009-0002-9186-8069}
\affiliation{
    \institution{Beijing University of Posts and Telecommunications}
    \city{Beijing} \country{China}
}

\author{Yongqian Sun} \orcid{0000-0003-0266-7899}
\affiliation{
    \institution{Nankai University \\ \& TKL-SEHCI}
    \city{Tianjin} \country{China}
}
\authornote{HL-IT, TKL-SEHCI, and BNRist are short for Haihe Laboratory of Information Technology
Application Innovation, Tianjin Key Laboratory of Software Experience and
Human Computer Interaction, and Beijing National Research Center for Information
Science and Technology, respectively.}
\author{Shenglin Zhang} \orcid{0000-0003-0330-0028}
\affiliation{
    \institution{Nankai University \\ \& HL-IT}
    \city{Tianjin} \country{China}
}

\author{Kun Wang} \orcid{0009-0000-1971-0860}
\affiliation{
    \institution{Tsinghua University \& BNRist}
    \city{Beijing} \country{China}
}

\author{Haiming Zhang} \orcid{0000-0001-9351-3320}
\author{Jianhui Li} \orcid{0009-0001-6253-9808}
\affiliation{
    \institution{CNIC, CAS}
    \city{Beijing} \country{China}
}

\author{Gaogang Xie} \orcid{0000-0003-4964-1135}
\affiliation{
    \institution{CNIC, CAS}
    \city{Beijing} \country{China}
}

\author{Xidao Wen} \orcid{0009-0002-4202-3748}
\affiliation{
    \institution{BizSeer}
    \city{Beijing} \country{China}
}

\author{Xiaohui Nie} \orcid{0000-0002-0371-854X}
\affiliation{
    \institution{CNIC, CAS}
    \city{Beijing} \country{China}
}

\author{Minghua Ma} \orcid{0000-0002-6303-1731}
\affiliation{
    \institution{Microsoft}
    \city{Redmond}
    \country{USA}
}

\author{Dan Pei} \orcid{0000-0002-5113-838X}
\authornote{Dan Pei is the corresponding author.}
\affiliation{
    \institution{Tsinghua University \& BNRist}
    \city{Beijing} \country{China}
}


\renewcommand{\shortauthors}{Y. Liu, C. Pei, L. Xu, B. Chen, M. Sun, Z. Zhang, Y. Sun, S. Zhang, K. Wang, H. Zhang, J. Li, G. Xie, X. Wen, X. Nie, M. Ma, D. Pei}

\renewcommand{\shortauthors}{Liu et al.}


\begin{abstract}
\label{sec:abs}
In recent decades, the field of software engineering has driven the rapid evolution of Information Technology (IT) systems, including advances in cloud computing, 5G networks, and financial information platforms. Ensuring the stability, reliability, and robustness of these complex IT systems has emerged as a critical challenge.
Large language models (LLMs) that have exhibited remarkable capabilities in NLP-related tasks are showing great potential in AIOps, such as root cause analysis of failures, generation of operations and maintenance scripts, and summarizing of alert information.
Unlike knowledge in general corpora, knowledge of Ops varies with the different IT systems, encompassing various private sub-domain knowledge, sensitive to prompt engineering due to various sub-domains, and containing numerous terminologies. Existing NLP-related benchmarks can not guide the selection of suitable LLMs for Ops (OpsLLM), and current metrics (e.g., BLEU, ROUGE) can not adequately reflect the question-answering (QA) effectiveness in the Ops domain. We propose a comprehensive benchmark suite, \textbf{OpsEval}, including an Ops-oriented evaluation dataset, an Ops evaluation benchmark, and a specially designed Ops QA evaluation method. Our dataset contains \mccount multiple-choice questions and \qacount QA questions. We have carefully selected and released 20\% of the dataset written by domain experts in various sub-domains to assist current researchers in preliminary evaluations of OpsLLMs\footnote{Data page is available at \url{https://github.com/NetManAIOps/OpsEval-Datasets}}. We test over \modelcount latest LLMs under various settings such as self-consistency, chain-of-thought, and in-context learning, revealing findings when applying LLMs to Ops. We also propose an evaluation method for QA in Ops, which has a coefficient of 0.9185 with human experts and is improved by 0.4471 and 1.366 compared to BLEU and ROUGE, respectively. 
Over the past one year, our dataset and leaderboard have been continuously updated. 
\end{abstract}




\received{2025-01-22}
\received[accepted]{2025-03-25}

\maketitle
\sloppy
\section{Introduction}
\label{sec:introduction}

IT Operations (Ops) plays a pivotal role in ensuring efficient and reliable functioning of software systems, including cloud computing, 5G networks, and financial platforms. With the rapid expansion of the Internet, the scale and complexity of software systems have grown exponentially, making traditional operations increasingly challenging. To address these challenges, artificial intelligence-assisted operations have emerged as a transformative approach, often referred to as ``AIOps'' by Gartner~\citep{lerner2017aiops}. AIOps uses artificial intelligence to tackle critical software engineering tasks such as anomaly detection, fault diagnosis, and performance optimization. 

In parallel, recent advances in large language models (LLMs) have further expanded the potential of intelligent solutions in software operations.
The latest models, such as
GPT-4o~\citep{openai2024gpt4o},
GPT-4V~\citep{openai2023gpt4v}, 
Meta-Llama-3~\citep{llama3modelcard}, and GLM-4~\citep{zeng2022glm}, 
have demonstrated exceptional generalization and task planning capabilities. As a result, these models have provided numerous opportunities to enhance downstream domain-specific applications. 
With its advanced 
text generation ability, 
LLM is well suited for Ops on tasks like question answering, information summarizing, and report analysis. Hereinafter, we refer to the LLM used for Ops as \textbf{OpsLLM}, regardless of whether they have been optimized specifically for Ops. 

While there are benchmarks for assessing general-purpose NLP-related capabilities, 
no benchmark exists to evaluate the effectiveness of LLMs or OpsLLMs in Ops tasks. There is an urgent need for an Ops benchmark that informs us about the performance of current LLMs on Ops tasks. On the other hand, a good benchmark can significantly aid the optimization process of OpsLLMs tailored for the Ops domain.
Nevertheless, due to the specialty of the Ops tasks, constructing an Ops benchmark presents the following challenges:

\begin{figure*}
    \centering
    \includegraphics[width=0.9\linewidth]{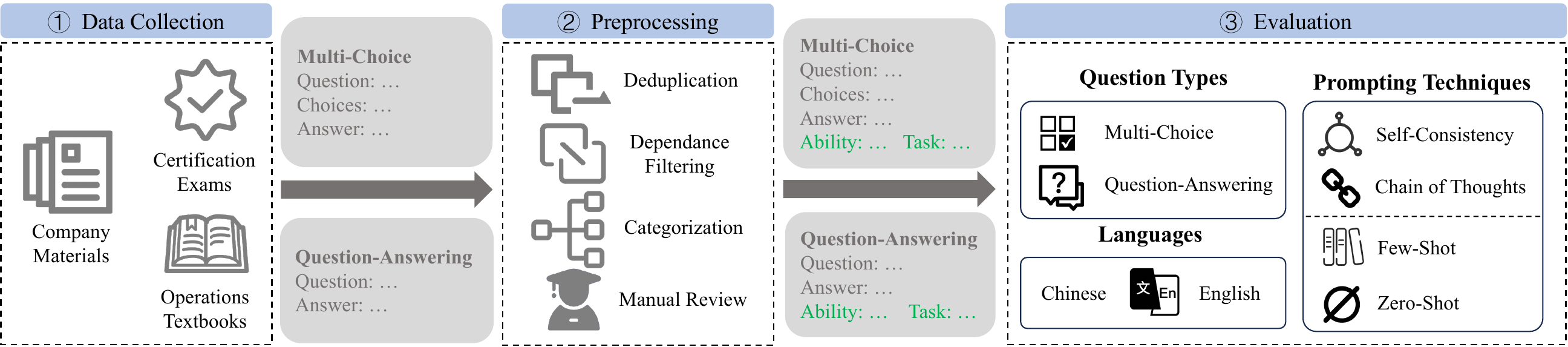}
    \caption{The framework of OpsEval.}
    \label{fig:framework}
    \Description[The framework of OpsEval.]{The framework of OpsEval.}
\end{figure*}

\begin{itemize}
    \item 
    \textbf{Sensitive data.} The Ops data is primarily sensitive and proprietary to companies, with very few publicly available data, making it difficult for any company to independently provide sufficient evaluation data to ensure confidence in the test results.
    \item 
    \textbf{Sub-domains.} The Ops field spans many sub-domains, like 5G communications, cloud computing, and bank transactions, each requiring a mix of capabilities, or “tasks,” such as network configuration or terminology explanation. The sheer number of sub-domains and tasks, combined with the absence of a systematic taxonomy, makes classifying questions challenging.
    \item 
    \textbf{Prompt sensitivity.} Due to the relatively proprietary nature of the Ops, existing LLMs have not undergone specialized supervised fine-tuning (SFT) for instruct following within the Ops field, the evaluation results are more sensitive to prompt engineering. Designing appropriate prompts for robust and accurate evaluation is challenging.
    \item 
    \textbf{QA metric.} Existing metrics like BLEU focus on linguistic similarity between model output and reference answers, which often fails to capture true performance in Ops tasks. In Ops, it’s essential to assess whether the model’s answers address key points in the reference and are supported by sufficient evidence, reflecting the precise meanings of domain-specific terms.
\end{itemize}

To address these issues, we propose \textbf{OpsEval}, a comprehensive benchmark suite for evaluating LLMs’ capability in the IT Operations domain, 
focusing on tasks essential to maintaining and troubleshooting live systems in real-world production environments. 
First, to tackle the challenge of benchmark data mostly being private, we initiated an AIOps community, which has attracted dozens of companies to participate. We have selected 9 representative sub-domains from the community, allowing continuous data contributions from community members. We aggregate data under the same sub-domain to ensure robustness in evaluation. Additionally, we generated multi-choice (MC) and question-answering (QA) questions as supplements based on publicly available network management books. 
To address the challenge of classifying the numerous sub-domains and tasks in the Ops field, we employ model-based pre-clustering and manual review to annotate eight tasks and three abilities.
Considering the prompt sensitivity of benchmark results, we systematically test model performance under self-consistency (SC), chain-of-thought (CoT), and few-shot in-context learning (ICL). 
Lastly, to address the inaccuracy of existing metrics in Ops QA evaluation, we design FAE-Score, which evaluates model responses based on fluency, accuracy, and evidence, with each criterion having its own dedicated assessment method.

The contributions of our paper are as follows:
\begin{enumerate}
    \item 
     We introduce \textbf{OpsEval}, the first bilingual multi-task dataset in the Ops domain, covering 8 tasks and 3 abilities with \allquestioncount questions. To assist researchers in preliminary evaluating their OpsLLMs, we have carefully selected and released 20\% of QAs from our benchmark licensed under CC-BY-NC-4.0, with the remaining 80\% of undisclosed data preventing unfair evaluations due to data leakage~\citep{wei2023skywork}.
    \item 
    Based on the dataset, we introduce the OpsEval evaluation benchmark, conducting independent and robust evaluations with various prompting techniques and a specifically designed evaluation metric, FAE-Score. Compared to the commonly employed BLEU and ROUGE metrics, FAE-Score exhibits a more pronounced congruence with the evaluations of human experts. Specifically, FAE-Score attains a correlation coefficient 0.9175 with expert assessments, surpassing the coefficients of 0.6705 for BLEU and -0.3957 for ROUGE. \item 
    Based on the results of OpsEval evaluation, we provide key observations and practical lessons to help domain practitioners make decisions such as whether existing models are sufficiently applicable within a specific sub-domain, the necessity for fine-tuning and whether model quantization compromises the effectiveness. 
\end{enumerate}
\section{OpsEval Benchmark}
\label{sec:benchmark}

Figure \ref{fig:framework} shows the overall framework of OpsEval from construction to evaluation. 
We collected data from multiple sources and then preprocessed it to enhance its quality.
Finally, we evaluated LLMs on the dataset using various prompt engineering techniques.

\subsection{Data Collection}
\label{sec:data-collection}

\begin{table}[t]
    \centering
    \caption{Overview of the question distribution in OpsEval by sub-domains, tasks and abilities.}
    \begin{subtable}[t]{0.45\textwidth}
        \centering
    \caption{The number of questions in OpsEval, grouped by their sub-domains.}
    \label{tab:stats}
    \footnotesize
    \begin{tabular}{llll}
    \toprule
        \textbf{Sub-domain} & \textbf{Source} & \textbf{Type} & \textbf{Questions}  \\
    \midrule
        Wired Network & Operation Textbooks & MC & 3901  \\
        \multirow{2}{*}{5G Communication} & \multirow{2}{*}{Certification Exams}  & MC & 2615  \\
        & & QA & 1162  \\
        Oracle Database & Company Materials  & MC & 497  \\
        Log Analysis & Company Materials  & QA & 420  \\
        DevOps & Company Materials  & QA & 154  \\
        Private Cloud & Company Materials & QA & 150 \\
        Securities Info. & Company Materials  & MC & 91\\
        Hybrid Cloud & Company Materials  & MC & 40\\
        Financial IT & Company Materials  & MC & 40 \\
    \midrule
        Total & & & \allquestioncount \\
    \bottomrule
    \end{tabular}
    \end{subtable}
    \hfill
    \begin{subtable}[t]{0.45\textwidth}
        \footnotesize
    \centering
    \caption{The distribution of different tasks and abilities of questions in OpsEval.}
    \label{tab:cat_dist}
    \begin{tabular}{clc}
    \toprule
    &\textbf{Category} & \textbf{Percentage} (\%)\\
    \midrule
    \multirow{8}{*}{\textbf{Task}}&Automation Scripts & $3.3$\\
    &Monitoring and Alerting & $5.2$\\
    &Performance Optimization & $5.3$\\
    &Software Deployment & $7.9$\\
    &Fault Analysis and Diagnostics & $13.7$\\
    &Network Configuration & $29.0$\\
    &General Ops Knowledge & $20.2$\\
    &Miscellaneous & $15.5$\\ 
    \midrule
    \multirow{3}{*}{\textbf{Ability}}&Knowledge Recall & $49.8$\\  
    &Analytical Thinking & $39.9$\\
    &Practical Application & $10.2$\\
    \bottomrule
    \end{tabular}
    \end{subtable}
\end{table}

\begin{table}[t]
\caption{Information of companies collaborating in OpsEval.}
\label{tab:companies}
\footnotesize
\centering
\begin{tabular}{lll}
\toprule
Organization    & Domain               & URL                                                  \\
\midrule
BOSC & Financial IT         & https://www.bosc.cn/zh/                              \\
Bizseer          & Ops service provider & https://www.bizseer.com/                             \\
ChinaEtek        & Internet             & https://www.ce-service.com.cn/                       \\
Data Foundation  & Internet             & https://www.dfcdata.com.cn/                          \\
Guotai Junan     & Securities           & https://www.gtja.com/                                \\
Huawei           & Communication        & https://www.huawei.com/ \\
Lenovo           & Hybrid Cloud         & https://www.lenovo.com/ \\
Rizhiyi          & Log Analysis         & https://www.rizhiyi.com/                             \\
ZTE              & Communication        & https://www.zte.com.cn/china/                        \\
Zabbix           & Ops service provider & https://www.zabbix.com/                              \\
Inspur & Ops service provider  & https://www.inspur.com/ \\
\midrule
Total            & 11                   &             \\
\bottomrule
\end{tabular}
\end{table}

Our benchmark questions have been collected from various sources; we summarize them into four categories: company materials, certification exams and Ops textbooks. Each source is highly esteemed globally and reviewed by our Ops collaborators.

\textbf{Company Materials.} include production environment materials like Ops tickets and error logs
, as well as internal documents and tests for Ops staff training. 
We have established cooperative relationships with \companycount companies, covering various sectors like telecommunications, finance, and Ops service/tool providers, and received expert collaboration and Ops materials from them.

Table \ref{tab:companies} shows the companies participating in the creation of OpsEval benchmark suite. Their industries include the Internet, telecommunications, cloud computing, finance, and securities, and each company has dispatched at least two experts to participate in the OpsEval work.

\textbf{Certification Exams.} include knowledge assessments necessary for becoming an Ops staff and are naturally in the form of multiple-choice and question-answering questions. We obtained the relevant study guidebooks for these certification exams from public book websites and extracted sample questions from them as one of the sources for Ops questions.

\textbf{Operations Textbooks}. We first constructed a seeding keyword list for the Ops field and searched for related books. The textbooks contain relatively complete knowledge content, which can provide experts with materials for question creation, and some books themselves also include a certain number of exercises at the end of the chapters.

\subsection{Preprocessing}
\label{sec:quality_enhancement}

We systematically carried out the preprocessing of our original data in the following stages:

\textbf{Deduplication:} Any repeated or highly similar questions are identified and removed to avoid redundancy in the test set. We calculate the cosine similarity of the question stems by bge-large-zh-v1.5 ~\citep{bge_embedding} to detect duplicate questions and identify pairs of questions with a similarity above a certain threshold (th=0.7). 

\textbf{Dependance Filtering:} 
We have filtered out questions that rely on external images or document content to ensure the completeness of the question content itself. The filtering process was done by two parallel lists of empirical keywords in the question stems and the responses of GPT-3.5-turbo. 
The keyword list is listed below.

\begin{insight}
question\_keywords = [`the figure', `the scenario', `the previous question']

fail\_pred\_keywords = [`unclear', `scenario is not provided', `cannot be determined', `none of the options', `none of the given options']
\end{insight}

\textbf{Question Categorization: }
We devise a categorization that captures many tasks that professionals confront in practical applications. The categorization process consists of two steps: automated screening and manual review.
We first use GPT-4 for topic modeling to gain rough insights about the dataset
and determine the relevance of each question to Ops, which resulted in more than 20 tasks but had an imbalanced distribution.
We then involved dozens of experts during the manual review process to categorize the questions into eight tasks and three abilities. 

\textbf{Tasks.} The details of each task are as follows.
\textit{General Knowledge} pertains to foundational concepts and universal practices within the Ops domain.
\textit{Fault Analysis and Diagnostics} focuses on detecting and addressing discrepancies or faults within a network or system, and deducing the primary causes behind those disruptions.
\textit{Network Configuration} revolves around suggesting optimal configurations for network devices like routers, switches, and firewalls to ensure their efficient and secure operations.
\textit{Software Deployment} deals with the deployment and management of software applications throughout the network or system.
\textit{Monitoring and Alerts} harnesses monitoring tools to supervise network and system efficiency and implements alert mechanisms to notify administrators of emerging issues.
\textit{Performance Optimization} is centered on refining the network and system for peak performance and recognizing potential enhancement areas.
\textit{Automation Scripts} involves the formulation of automation scripts to facilitate processes and decrease manual intervention for administrators.
\textit{Miscellaneous} comprises tasks that do not strictly adhere to the aforementioned classifications or involve a combination of various tasks.

\textbf{Abilities.} Different questions require different levels of ability to answer. 
\textit{Knowledge Recall} primarily test a model's capacity to recognize and recall core concepts and foundational knowledge, which are akin to situations where professionals identifies a standard procedure or recognize an issue based solely on previous knowledge.
\textit{Analytical thinking} necessitates a deeper level of thought, expecting the model to dissect a problem, correlate diverse pieces of information, and derive a coherent conclusion. It mirrors scenarios where professionals troubleshoot complex issues by leveraging their comprehensive understanding.
\textit{Practical Application} challenges a model to apply its knowledge or analytical conclusions to provide actionable recommendations for specific scenarios. It mirrors situations where professionals make decisions or suggest solutions based on in-depth analysis and expertise.

\begin{figure}
    \centering
    \includegraphics[width=0.8\linewidth]{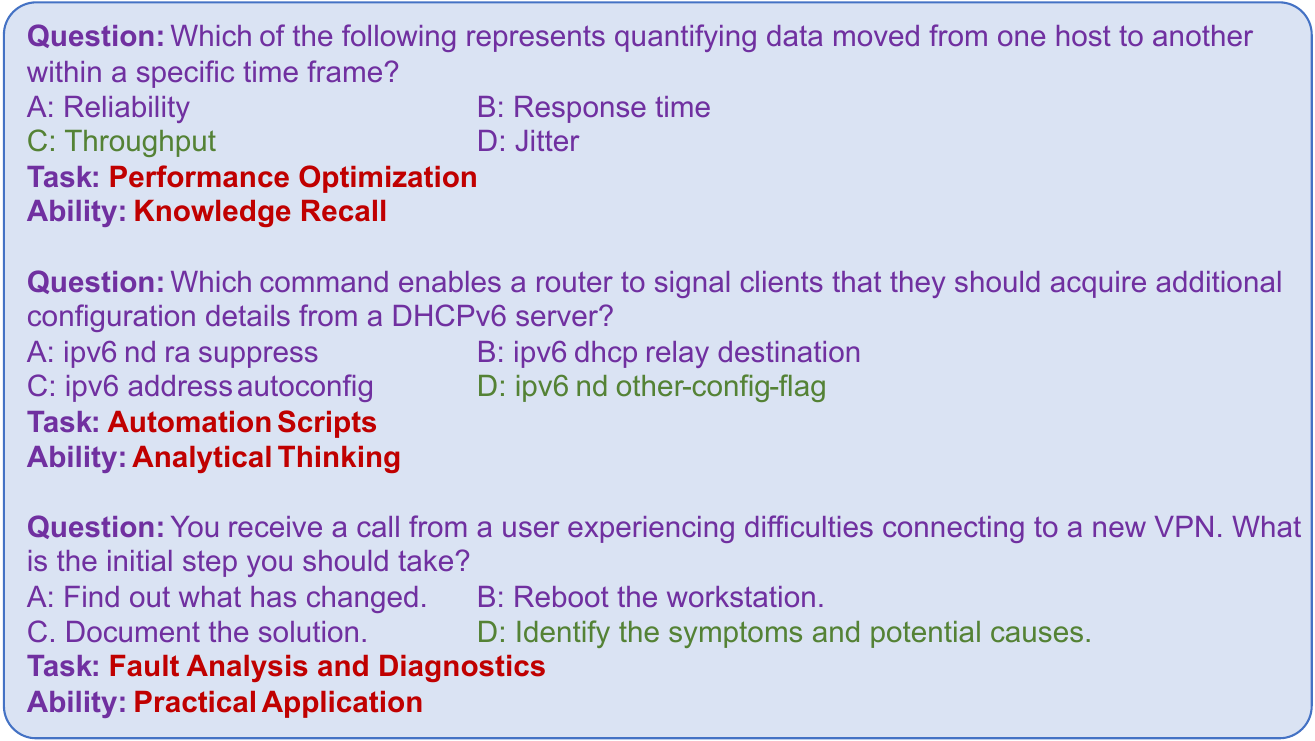}
    \caption{Three examples of the processed questions.}
    \label{fig:format_example}
    \Description[Three examples of the processed questions.]{Three examples of the processed questions.}
\end{figure}

Figure \ref{fig:format_example} illustrates examples in our question set, demonstrating our classification methodology. The distribution of the questions across the tasks and ability levels is shown in Table \ref{tab:cat_dist}.

\textbf{Manual Review:} In the manual review step, we asked Ops experts from the industry to inspect the results of the previous three automated steps, including confirming duplicate and invalid questions and examining the classification results of GPT-4. In our work, an expert is defined as an individual with ten or more years of professional experience in their field, whether as an employee or a researcher.
Experts were also asked to drop the questions unrelated to the Ops field. 
We split the dataset by n-folds and ensure each fold has at least two experts to review. The review process followed a standardized annotation guideline, which is available in our dataset repository. 
As listed in Table \ref{tab:stats}, this quality enhancement process resulted in a refined test set of approximately 7,000 multi-choice and 2,000 question-answering questions.

\begin{figure*}[t]
    \centering
    \includegraphics[width=0.85\linewidth]{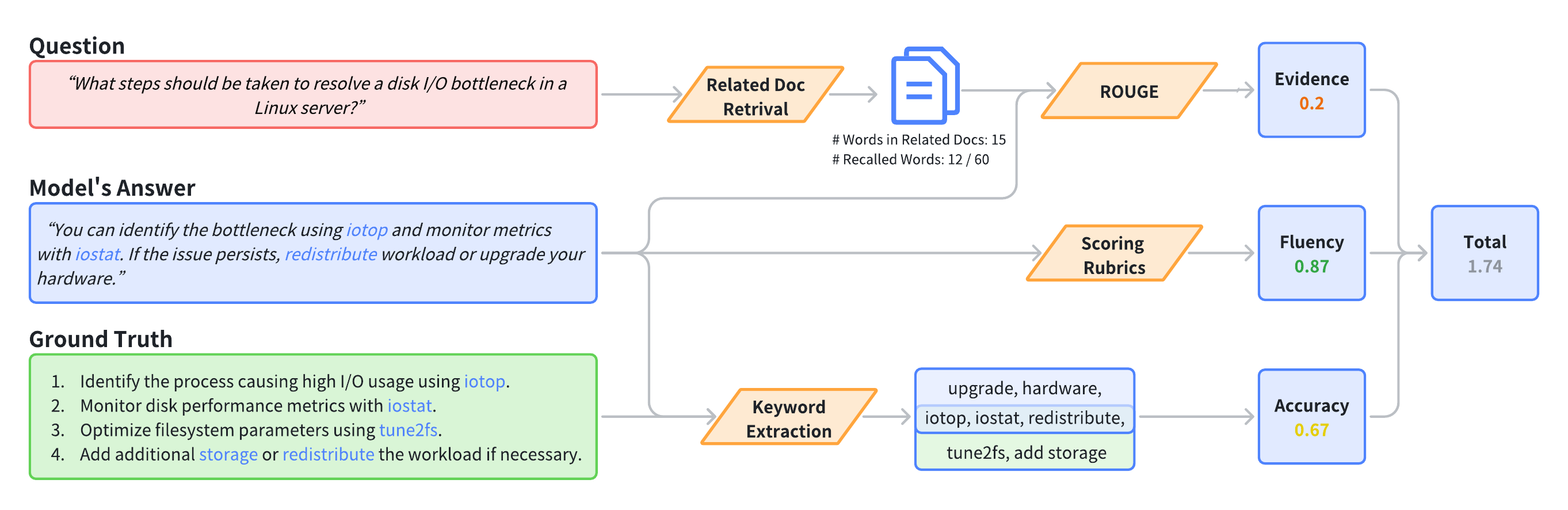}
    \caption{The FAE-Score pipeline.}
    \label{fig:fae-score-pipeline}
    \Description[The FAE-Score pipeline.]{The FAE-Score pipeline.}
\end{figure*}

\subsection{Evaluation Settings}
\label{sec:evaluation_settings}

\textbf{Multi-choice questions} offer a structured approach with definitive answers. These questions are straightforward and provide a clear metric for assessment. 
We use \textbf{accuracy} as the metric. 
A choice-extracting function based on regular expressions is used to extract the predicted answer of LLMs. Then, we calculate the accuracy based on the extracted answer and the ground-truth labels.

\begin{table}
\caption{Models evaluated in this paper.}
\footnotesize
\centering
\begin{tabular}{lllll}
    \toprule
        {\bf Model}             
        & {\bf \#Parameters}    & {\bf Access}  & \textbf{License} \\ 
        
        \midrule
        
        GPT-4/3.5-turbo                  
        & \textit{undisclosed}      & API    & Proprietary   \\
        
        ERNIE-Bot-4.0 
        & \textit{undisclosed} & API& Proprietary\\
        GLM4/GLM3-turbo 
        & \textit{undisclosed} & API& Proprietary\\
        Meta-LLaMA-3        
        & 8B & Weights 
        & Llama 3 Community \\
        LLaMA-2             
        & 7/13/70B                       & Weights  
        & Llama 2 Community \\
        Qwen-Chat
        & 7/14/72B                        & Weights 
        & Qianwen LICENSE \\
        Qwen1.5-Chat
        & 14B & Weights 
        & Qianwen LICENSE \\
        InternLM2-Chat
        & 7/20B                        & Weights & Apache-2.0  \\
        DevOps-Model   
        & 14B & Weights& Apache-2.0 \\
        Baichuan2-Chat      
        & 13B                       & Weights & Apache-2.0 \\
        ChatGLM3             
        & 6B                        & Weights & Apache-2.0   \\
        Mistral 
        & 7B & Weights & Apache-2.0 \\
        Gemma 
        & 2/7B & Weights & Gemma license\\
        Claude-3-Opus 
        & \textit{undisclosed} & API & Proprietary \\
        Qwen2-Instruct 
        & 7/72B & Weights 
        & Qianwen LICENSE \\
    \bottomrule
\end{tabular}
\label{tab:model-overview}
\end{table}

\textbf{Question-answering questions} are evaluated using a metric designed specifically for OpsEval, called \textbf{FAE-Score}, which is explained in detail in the subsequent section. Additionally, we perform expert evaluations and calculate BLEU \citep{papineni2002bleu}, ROUGE \citep{lin2004rouge} and RAGAS \citep{es-etal-2024-ragas} scores for comparison purposes, as reference to validate the accuracy of FAE-Score. 

We use the same three criteria to evaluate the responses of various models for both FAE-Score and Expert Evaluation:

\begin{itemize}
    \item \textbf{Fluency}. Assessment of the linguistic fluency in the model's output and compliance with the question-answering question's answering requirements.
    \item \textbf{Accuracy}. Evaluation of the precision and correctness of the model's output, including whether it adequately covers key points of the ground-truth answer.
    \item \textbf{Evidence}. Examine whether the model's output contains sufficient argumentation and evidential support to ensure the credibility and reliability of the answer.
\end{itemize}

In Expert Evaluation, we asked experts to score it between 0 and 3 for each criterion. During the scoring, the raw question, the detailed answer and its key points, and the output of an anonymous model are given at each iteration.

\textbf{Prompting Techniques.}
We use various settings to evaluate LLMs on OpsEval to get a comprehensive overview of their performance. We evaluate LLMs in zero and few-shot (3-shot) settings. 
For each setting, we evaluate LLMs in four sub-settings of prompt engineering, that is, naive answers (Naive), self-consistency (SC)~\citep{wang2023selfconsistency}, chain-of-thought (CoT)~\citep{wei2023chainofthought}, self-consistency with chain-of-thought (CoT+SC). We set the number of queries in SC to \textbf{5}.

\textbf{Models.}
\label{sec:models}
We evaluate popular LLMs covering different weights from different organizations. The model selection was guided by specific criteria: We aimed to include the latest and most advanced large language models, with a particular focus on those capable of handling Chinese input.
The detailed information of all \modelcount LLMs can be found in Table \ref{tab:model-overview}.

\subsection{FAE-Score}

Figure \ref{fig:fae-score-pipeline} shows the basic pipeline of our designed QA metric, FAE-Score. Here, we elaborate each evaluation methodology of each criterion.

\begin{figure}
    \centering
    \includegraphics[width=0.4\textwidth]{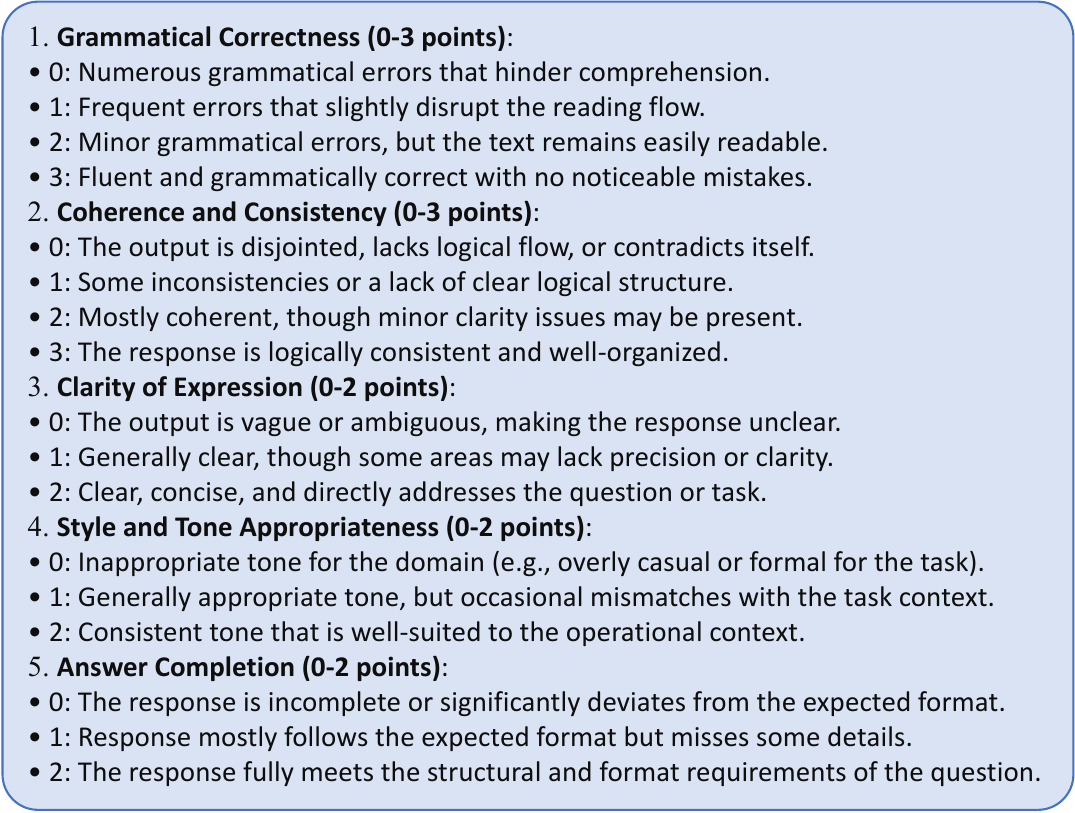}
    \caption{Scoring rubrics for Fluency metric.}
    \label{fig:fluency-rubrics}
    \Description[Scoring rubrics for Fluency metric.]{Scoring rubrics for Fluency metric.}
\end{figure}

\textbf{Fluency.} In Ops settings, the fluency of a model’s output is crucial because the results are intended for human consumption by technical personnel. Unlike other generic benchmarks, the tasks in the Ops domain require clear and unambiguous communication, as the model’s outputs may guide decision-making in production scenarios. 
To evaluate fluency in model outputs, we adapted the scoring rubrics methodology mentioned in \citet{kim2024prometheus}. We use Qwen2-72B-Instruct as the evaluation model, for its strong performance in general language generation \citep{2023qwen} and its consistent multilingual capabilities. We assess the fluency of various model outputs based on grammar, coherence, clarity, appropriateness of style, and answer completeness, as shown in Figure \ref{fig:fluency-rubrics}.

\textbf{Accuracy.} Traditional metrics such as BLEU and ROUGE fall short in this vertical domain because they often fail to capture the key factual content within long-form responses. This results in inflated scores due to irrelevant word matches, making these metrics insufficient for accuracy evaluation in the highly specialized and knowledge-driven Ops context. To address these shortcomings, we take inspiration from \citet{es-etal-2024-ragas}, using a keyword extraction method to evaluate the accuracy of model outputs. A judge model \citep{openai2023gpt4} is then employed to match the keywords from the model’s response with the keywords from the standard answer. The final accuracy score is calculated by determining the F1-Score, which balances precision and recall for the matched keywords.

\begin{equation}
    \textbf{Accuracy} = 2 \cdot \frac{\text{P} \cdot \text{R}}{\text{P} + \text{R}}
\end{equation}

\begin{equation}
    \text{P} = \frac{\# {\text{Matched Keywords}}}{\# {\text{Keywords in Model Output}}}
\end{equation}

\begin{equation}
    \text{R} = \frac{\# {\text{Matched Keywords}}}{\# {\text{Keywords in Ground Truth}}}
\end{equation}

\begin{figure*}[ht]
    \centering
    \includegraphics[width=0.88\linewidth]{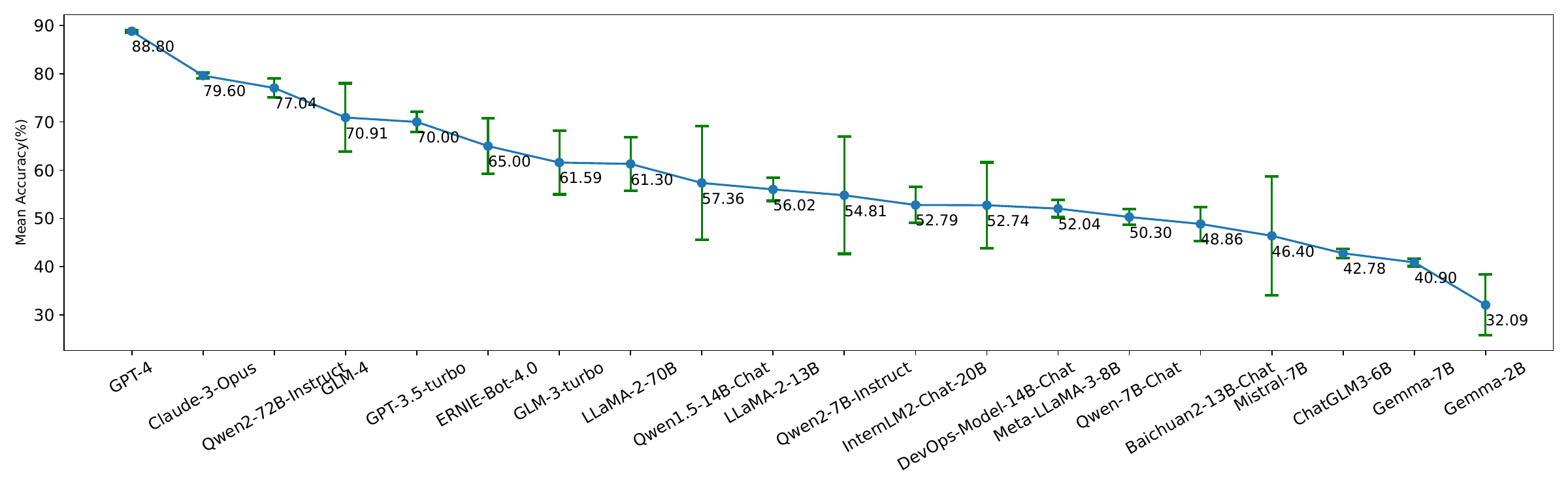}
    \caption{LLMs' overall performance on Wired Network Operations English test set (3-shot). \normalfont Models are ranked based on their mean accuracy among different settings. The error bars represent the variance in the model's accuracy across different prompting techniques.}
    \label{fig:network_eng_3shot}
    \Description[LLMs' overall performance on Wired Network Operations English test set (3-shot).]{LLMs' overall performance on Wired Network Operations English test set (3-shot). Models are ranked based on their mean accuracy among different settings. The error bars represent the variance in the model's accuracy across different prompting techniques.}
\end{figure*}

\textbf{Evidence.} Model responses must not only be accurate but also well-supported by relevant, authoritative information. To evaluate the evidence behind a model’s response, we implement a ROUGE-based method to measure the overlap between the generated output and the content of related documents retrieved through similarity search. We used bge-large-zh \citep{bge_embedding} for document embedding and FAISS \citep{douze2024faiss} for similarity search. By retrieving documents that closely match the question, we can assess whether the model’s response appropriately references or aligns with this external information. We use ROUGE, as a recall-oriented metric, captures how much of the content in the relevant documents is reflected in the model’s output. This ensures that the model does not simply generate plausible-sounding answers but grounds its responses in factual evidence from trusted sources.

\begin{equation}
\textbf{Evidence} = \text{ROUGE}_\text{Recall}(R, D) = \frac{\# \text{Overlapping Words}}{\# \text{Words in } D}
\end{equation}

\subsection{Open-source Policy}

We released 20\% of the OpsEval dataset to support research and community contributions. The subset was proportionally sampled across sources and sub-domains, with proprietary content reviewed to remove sensitive information.
This sample dataset offers a clear view of question types and topics in the benchmark, helping researchers grasp the evaluation scope. It also supports local model evaluation for quicker iteration and can seed automatic QA generation~\citep{selfinstruct}, enriching Ops-specific data for future development.
While this subset is available for users' self-evaluation, the complete dataset remains undisclosed. By ensuring that the test set answers are not leaked, we guarantee the reliability and non-leakage of the OpsEval benchmark. 
\section{Result Analysis}
\label{sec:evaluation}

\subsection{Overall Performance}

The results of the few-shot evaluation with four settings on the Wired Network Operation test set are shown 
in Figure \ref{fig:network_eng_3shot}. 
\footnote{For results of the other sub-domains and settings, please check our official leaderboard website \url{https://opseval.cstcloud.cn/content/leaderboard}.}
While closed source models like GPT-4 and Claude-3-Opus performs well on the OpsEval benchmark, open-sourced LLMs yield generally worse evaluation results than those in general domains like MMLU~\citep{MMLU} and CEval~\citep{C_Eval}. This comparison highlights the necessity of explicitly fine-tuning OpsLLM for the Ops field.
Recent open-sourced models like Qwen2-72B-Chat, exhibit competitive performance in multi-choice questions, thanks to their fine-tuning process and the quality of their training data. 
Furthermore, we observed significant variability in how different LLMs respond to various prompt engineering techniques. Given the critical importance of stability in the Ops domain, it is essential to consider a model's sensitivity to prompts when selecting foundation model. Further research into prompt engineering is needed to improve model performance and reliability in this domain.

\textbf{Observations:}
1) Few-shot and CoT can significantly increase performance if the model is tuned to adapt to these techniques, while SC may have little influence on highly consistent LLMs.
2) Smaller models with weaker abilities are less stable with advanced prompts. Simpler prompts work better for them.

\textbf{Pratical Lesson}:
The choice of fundamental models should be a balance between their performance (average score) and robustness (variance) under different prompt settings.

\subsection{Performance on Different Tasks and Abilities}

\begin{figure*}[t]
  \centering
  \includegraphics[width=0.75\linewidth]{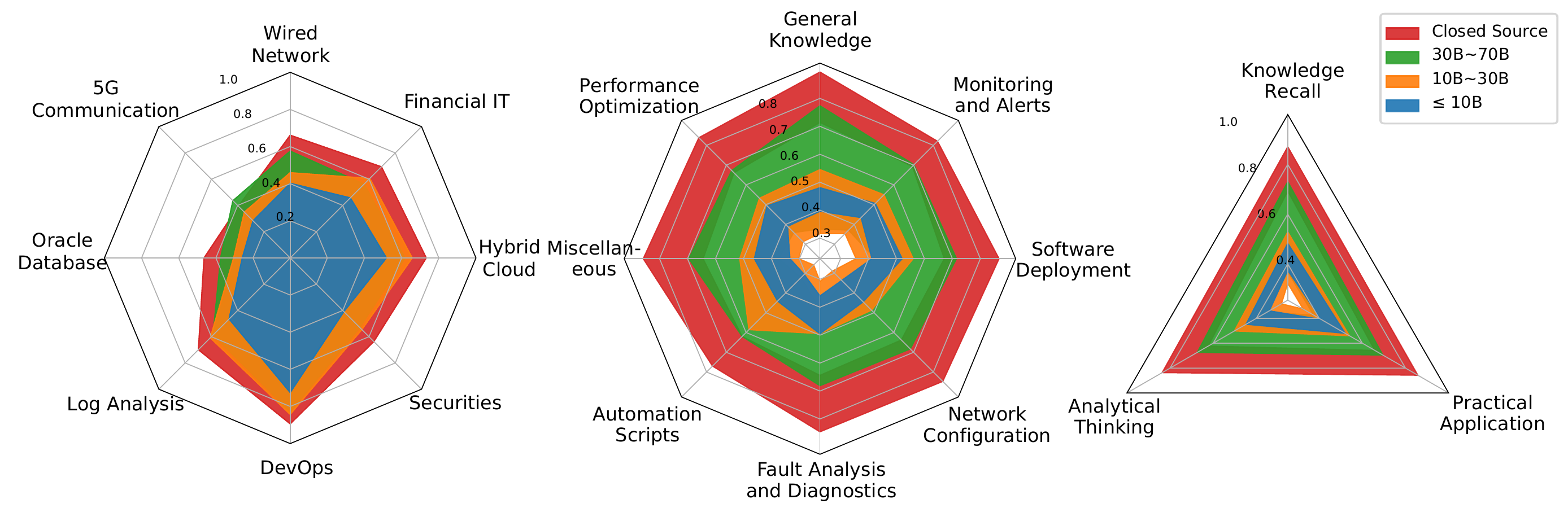}
  \caption{LLMs' performance on eight Ops sub-domains, eight tasks and three abilities. Each colored area presents the lower and upper bound of the corresponding parameter-size group.}
  \label{fig:scene_ability_radar}
  \Description[LLMs' performance on eight Ops sub-domains, eight tasks and three abilities.]{LLMs' performance on eight Ops sub-domains, eight tasks and three abilities. Each colored area presents the lower and upper bound of the corresponding parameter-size group.}
\end{figure*}

\begin{figure*}[t]
    \centering
    \includegraphics[width=0.75\linewidth]{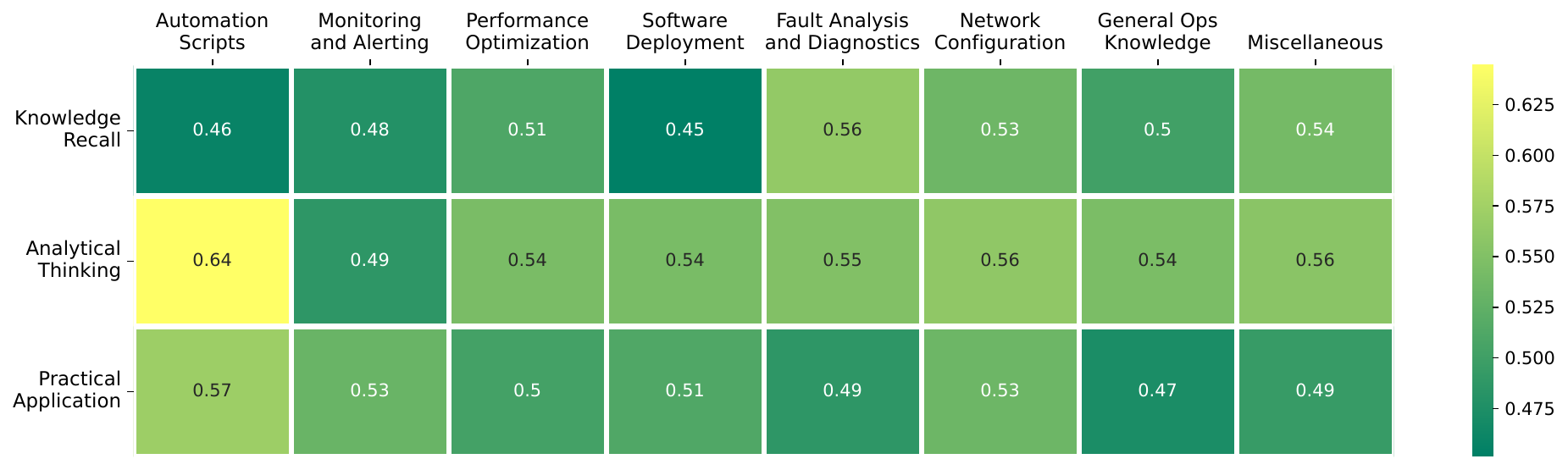}
    \caption{Heatmap of failure case distribution regarding tasks and abilities. The values represent the proportion of failure cases across all LLMs; yellower areas indicate higher failure rates.}
    \label{fig:fail_heatmap}
    \Description[Failure case distribution heatmap]{Heatmap of failure case distribution regarding tasks and abilities. The values represent the proportion of failure cases across all LLMs; yellower areas indicate higher failure rates.}
\end{figure*}

To investigate how LLMs perform in each Ops sub-domain and each task,  and to what extent they possess the general abilities,
we summarize the result of different parameter-size groups of LLM in Figure \ref{fig:scene_ability_radar}.
Regarding the eight tasks we tested, LLMs yield higher accuracy in General Knowledge tasks, while their performance drops and varies drastically in highly specialized tasks like Automation Scripts and Network Configuration, reflecting the impact of specialized corpus and domain knowledge on the performance of LLMs. 
By grouping LLMs by their parameter size, we find that while LLMs with 10B-30B parameters have higher accuracy in their best cases compared with LLMs with no more than 10B parameters, different 10B-20B LLMs' performance varies drastically. 
To provide systematic practical lessons for researchers in the operations domain on pre-training and fine-tuning OpsLLM, we have analyzed the error rates of LLMs across the 8 tasks and 3 abilities in Figure \ref{fig:fail_heatmap}. By examining the focus areas across different categories, we have identified key research targets for capability training.

\textbf{Observations:} Among the 24 categories of results, models performed the worst in Analytical Thinking for Automation Scripts. This indicates that current models can only recall the learned scripts but struggle to infer their logical relationships. Similarly, Analytical Thinking showed the lowest performance across the three major tasks, indicating that current OpsLLM models still have some way to go before becoming foundational models for Ops Agents. Thus, researchers should focus on inference-related SFT (supervised fine-tuning) datasets.

\textbf{Insights:}
1) Among different sub-domains of Ops, 5G communication and database demand further pretraining and fine-tuning.
2) To be capable of an Ops agent, the foundation model must be able to make a connection between specialized domain knowledge.

\subsection{Performance on Question-Answering}

\begin{table*}[t]
\caption{LLMs' performance on English network operations question-answering problems.}
\centering
\footnotesize{
\begin{tabular}{lccccccccccc}
    \toprule
    \multirow{2}{*}{Model} & \multirow{2}{*}{ROUGE(\%)} & \multirow{2}{*}{BLEU(\%)} & \multirow{2}{*}{RAGAS(0-10)} & \multicolumn{2}{c}{Fluency} & \multicolumn{2}{c}{Accuracy} & \multicolumn{2}{c}{Evidence}  & \multicolumn{2}{c}{FAE-Total}\\
    \cmidrule(lr){5-6} \cmidrule(lr){7-8} \cmidrule(lr){9-10} \cmidrule(lr){11-12}
    & & & & FAE & Expert & FAE & Expert & FAE & Expert & FAE & Expert\\
    \midrule

    GPT-3.5-turbo & 12.26 & 6.78 & 9.23 & 9.38 & 9.12 & 8.06 & 9.65 & 6.21 & 8.11 & 23.65  & 26.88 \\
    LLaMA-2-70B & 7.74 & 4.2 & 6.04 & 8.69 & 8.25 & 7.71 & 8.79 & 9.08 & 8.98 & 25.48  & 26.02 \\
    LLaMA-2-13B & 4.98 & 3.43 & 8.23 & 8.47 & 9.84 & 7.32 & 9.34 & 8.81 & 7.27 & 24.60  & 26.44 \\
    Chinese-Alpaca-2-13B & 3.25 & 1.85 & 5.32 & 5.53 & 8.05 & 6.99 & 7.95 & 6.23 & 6.23 & 18.75  & 22.24 \\ 
    Baichuan-13B-Chat & 4.76 & 0.35 & 7.93 & 7.16 & 7.98 & 8.71 & 7.84 & 6.66 & 7.31 & 22.53  & 23.13 \\
    Qwen-7B-Chat & 11.82 & 4.33 & 4.92 & 7.63 & 5.82 & 6.42 & 7.27 & 6.57 & 5.37 & 20.62  & 18.47 \\
    ChatGLM2-6B & 9.71 & 5.07 & 5.32 & 5.12 & 7.96 & 6.41 & 6.39 & 6.14 & 4.32 & 17.67  & 18.67 \\
    InternLM-7B & 13.27 & 0.54 & 6.21 & 4.99 & 5.16 & 5.00 & 4.90 & 4.75 & 4.28 & 14.74  & 15.77 \\
    Chinese-LLaMA-2-13B & 9.19 & 0.24 & 7.34 & 6.98 & 4.64 & 5.29 & 6.32 & 4.63 & 8.34 & 16.90  & 17.88 \\ 
    
    \bottomrule
\end{tabular}
}
\label{tab:eng_sub_score2}
\end{table*}

\begin{figure}
    \centering
    \includegraphics[width=\linewidth]{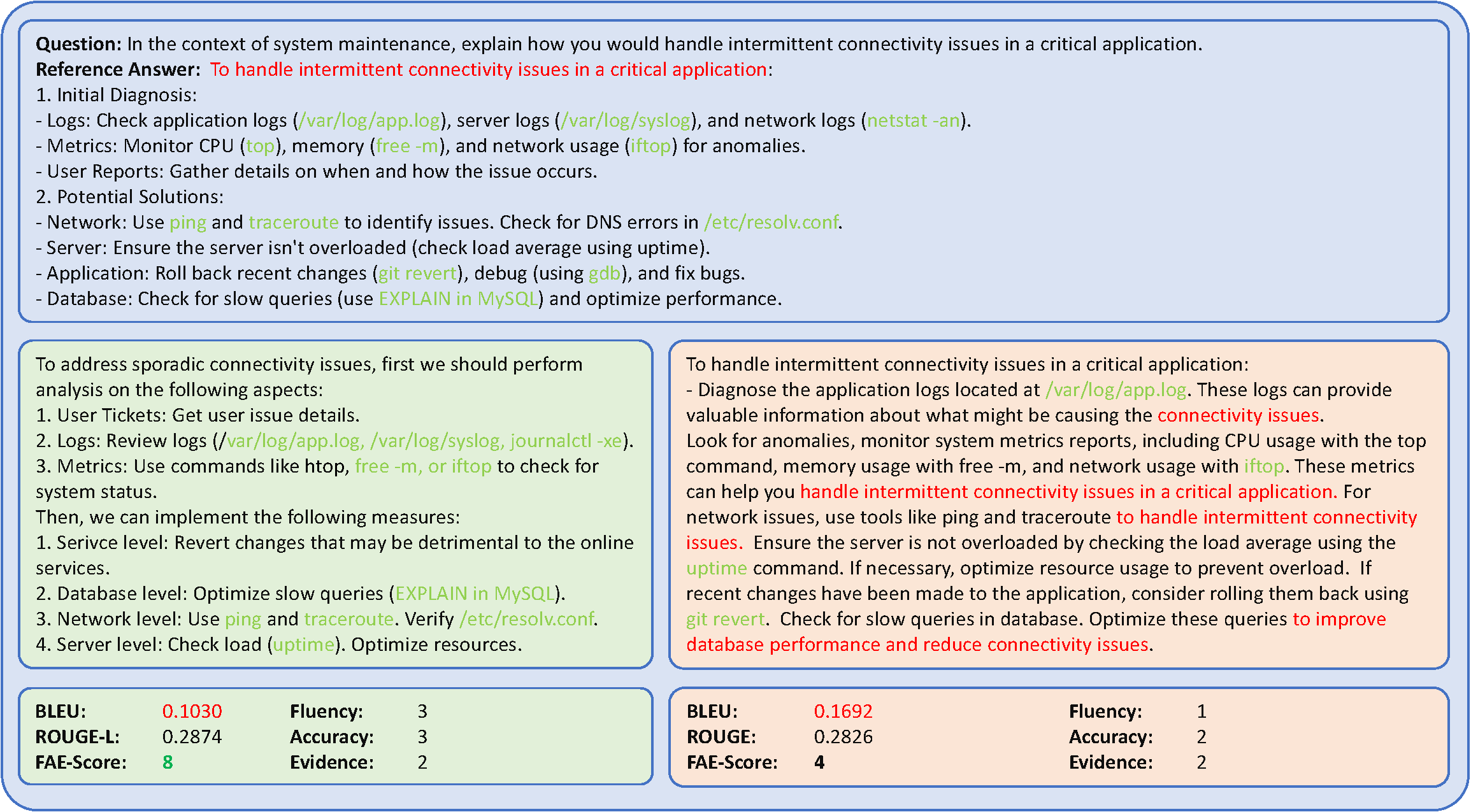}
    \caption{Case analysis on QA metrics.}
    \label{fig:qa_metrics_example}
    \Description[Case analysis on QA metrics.]{Case analysis on QA metrics.}
\end{figure}

Table \ref{tab:eng_sub_score2} presents the evaluation results of 200 question-answering English questions across four metrics: ROUGE, BLEU, RAGAS, FAE-Score, and Expert-Evaluation. 
To gain more insight into how different metrics perform in QA evaluation, we use Figure \ref{fig:qa_metrics_example} 
as a case analysis. While BLEU and ROUGE are efficient in natural language comparison, they lack semantic information to determine which part of the context is more important than others. Knowing that a given benchmark evaluates QA based on BLEU/ROUGE, there is an obvious way to trick the metric: repeat patterns occurring in the question, gaining a higher possibility to match some patterns in the reference answer. Due to their lack of semantic information related to Ops and the potential hack, traditional metrics like BLEU are unsuitable for specialized benchmarks. Instead, with specialized prompting and seperately designed methodology for each criterion (Fluency, Accuracy and Evidence), FAE-Score can comprehensively evaluate models' QA performance, with the Accuracy metric picking up those important keywords and not be influenced by repeated words that contain no useful information, and the Evidence metric checking the recall of relevant supporting contents.
In Section \ref{sec:validation}, we discuss the alignment between different metrics and expert evaluation, validating the effectiveness of FAE-Score in automated QA evaluation within the Ops domain.

\textbf{Insight:} In specialized domains, Ops specifically, traditional NLP metrics like BLEU and ROUGE cannot comprehend the key components in the reference answer, resulting their evaluation lacking practical significance. FAE-Score is suitable for large-scale qualitative evaluations in the Ops field.

\subsection{Performance on Different Quantization parameters}

\begin{figure*}
  \centering
  \includegraphics[width=0.8\linewidth]{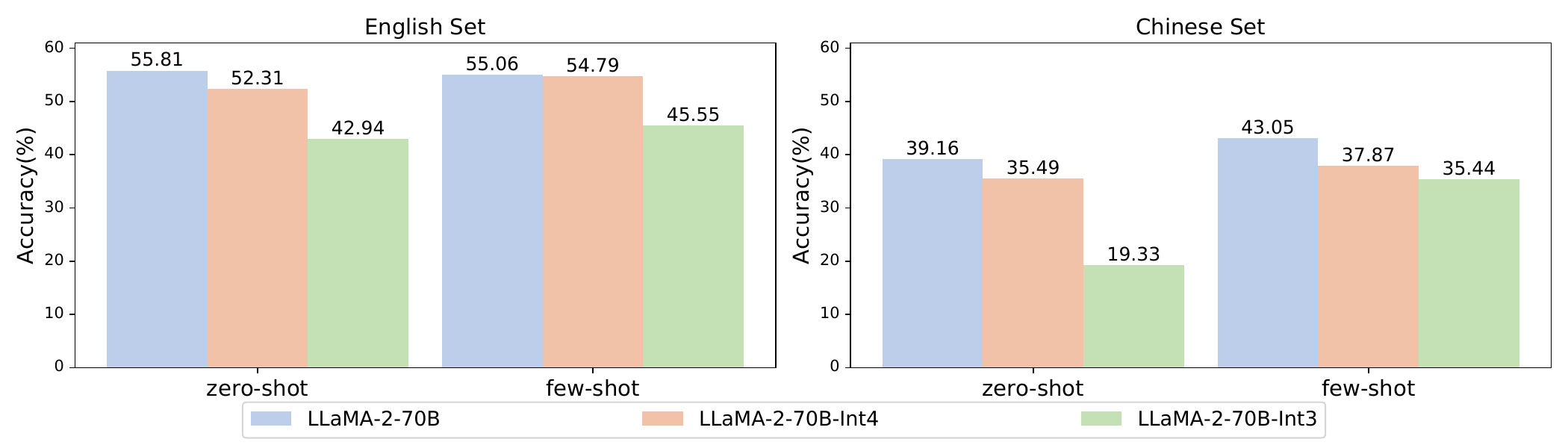}
  \caption{LLaMA-2-70B's performance of different quantization parameters. \normalfont Both zero-shot and few-shot evaluations have been conducted on Wired Network Operations test set under the naive setting.}
  \label{fig:quantization-res}
  \Description[LLaMA-2-70B's performance of different quantization parameters.]{LLaMA-2-70B's performance of different quantization parameters. \normalfont Both zero-shot and few-shot evaluations have been conducted on Wired Network Operations test set under the naive setting.}
\end{figure*}

We conducted experiments on different quantized versions of LLaMA-2-70B and obtained various results and conclusions. 
Figure~\ref{fig:quantization-res} shows the accuracy of LLaMA-2-70B of different quantization parameters on English and Chinese questions. We do both zero-shot and few-shot evaluation with the naive setting.

LLaMA2-70B-Int4 can achieve an accuracy close to LLaMA-2-70B without quantization. On English multi-choice questions, the accuracy of the GPTQ model with 4-bit quantization parameters is 3.50\% lower in zero-shot evaluation and 0.27\% in few-shot evaluation compared to LLaMA-2-70B. For Chinese questions, the accuracy of LLaMA2-70B-Int4 is 3.67\% lower in zero-shot evaluation and 5.18\% in few-shot evaluation compared to LLaMA-2-70B.
However, LLaMA2-70B-Int3 has a performance degradation that cannot be ignored. On average, the accuracy of LLaMA2-70B-Int3 in English set has a 12.46\% degradation compared to LLaMA-2-70B and a 9.30\% degradation compared to LLaMA2-70B-Int4. 
Overall, although the performance of the INT4 version decreases in both English and Chinese, the decline does not exceed 10\%. However, the performance drop in the INT3 version is more significant, requiring careful consideration in practical applications.

\textbf{Practical Lesson:} Quantization with more than 3 bits can effectively reduce computation and memory costs while preserving performance.

\section{Validation}
\label{sec:validation}

\begin{table}
    \centering
    \caption{Validation results.}
    \begin{subtable}[t]{0.5\textwidth}
        \centering
    \caption{Measurement of potential test data leakage.}
    \footnotesize
    \begin{tabular}{ccccc}
    \toprule
        Dataset & $L_{test}$ & $L_{ref}$ & $\Delta L$ & $\geq 0?$\\
    \midrule
    Alpaca	&1.9940	&2.3542&	\biased{-0.3602} & \xmark\\
    Alpaca-GPT4 &1.4988 	 &	1.7636& \biased{-0.3910} & \xmark \\
    \midrule
    CEval & 2.5708 & 2.3099 & \unbiased{0.2608} & \cmark \\
    MMLU & 2.5475 & 2.1898 & \unbiased{0.3577} & \cmark \\
    OpsEval & 2.9854 & 2.6280 & \unbiased{0.3050} & \cmark \\
    \bottomrule
    \end{tabular}
    \label{tab:corruption_result}
    \end{subtable}
    \begin{subtable}[t]{0.5\textwidth}
        \caption{Pearson correlation coefficients between Expert-Evaluation metrics and Automated metrics. Total is the sum of Fluency, Accuracy, and Evidence.}
    \footnotesize
    \centering
    \begin{tabular}{ccccc}
    \toprule
        Metric & Total & Flu. & Acc. & Evi. \\
    \midrule
        ROUGE & -0.44734 & -0.49207 & -0.40889 & -0.31821 \\
BLEU & 0.47139 & 0.46369 & 0.55330 & 0.05977 \\
RAGAS & 0.57169 & 0.40029 & 0.51151 & 0.41928 \\
FAE-Score & \textbf{0.91848} & \textbf{0.54757} & \textbf{0.81523} & \textbf{0.58160} \\
    \bottomrule
    \end{tabular}
    \label{tab:gpt4_humaneval_corr}
    \end{subtable}
\end{table}

\subsection{Benchmark Leakage Test}
\label{sec:leakage_test}

\begin{figure}
    \centering
    \includegraphics[width=\linewidth]{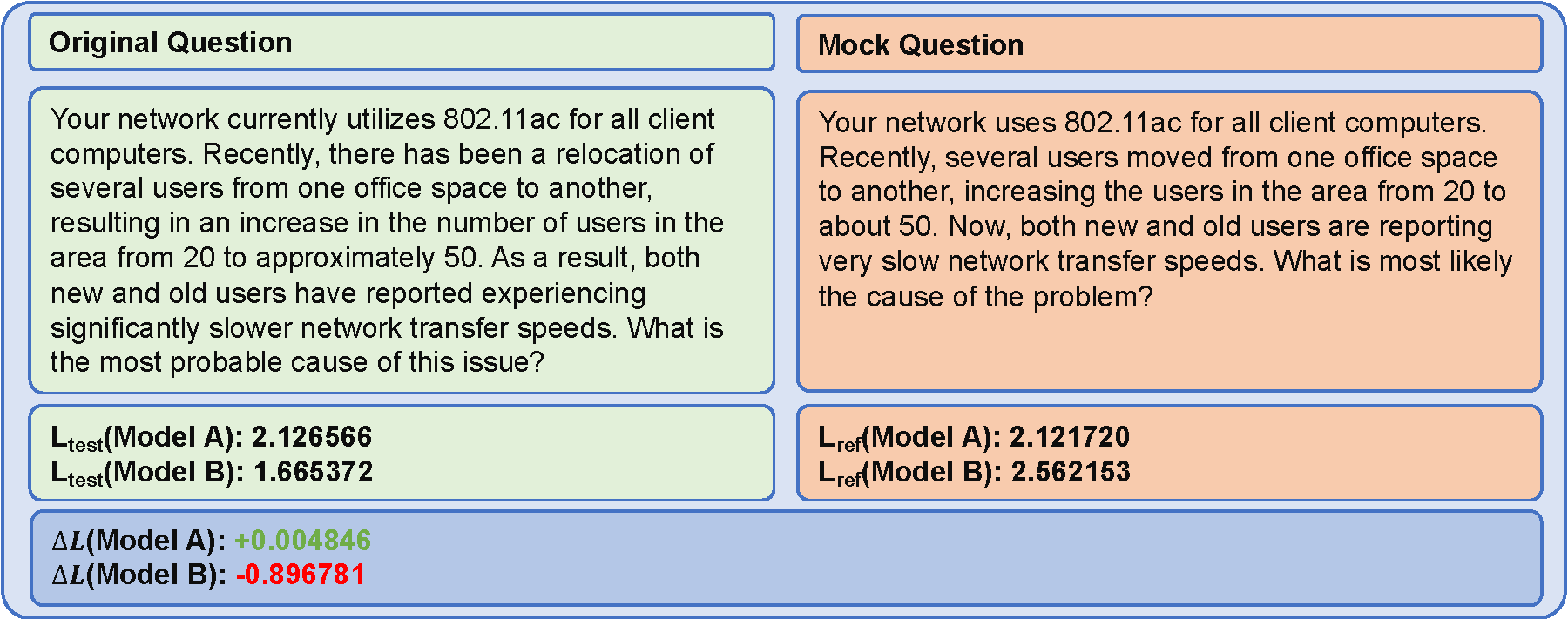}
    \caption{An example for leakage test. }
    \label{fig:leakage_test_example}
    \Description[An example for leakage test.]{An example for leakage test.}
\end{figure}

For the fairness of a benchmark suited for LLM, avoiding potential bias emerging from test set leakage is necessary.
We adapted the methodology from~\cite{wei2023skywork} to perform a leakage test on OpsEval's dataset. 
We evaluate the LLM loss on samples from different datasets for several LLMs and calculate the average loss. For each dataset, we compare LLM loss on
the test split ($L_{test}$) and a specially curated reference set ($L_{ref}$ ) generated
by GPT-4, designed to mimic the testing dataset. 
While ~\cite{wei2023skywork} only asked GPT-4 to generate similar questions to the GSM8K~\citep{cobbe2021gsm8k} dataset, we require GPT-4 to rewrite the question while preserving its original meaning.
We define a key metric: $\Delta L = L_{test} - L_{ref}$, with a threshold of $\Delta L < 0$ indicating potential test data leakage.
A negative $\Delta L$ suggests that the LLM's lower $L_{test}$ comes from overfitting the test set rather than understanding the questions, indicating potential leakage. 
Figure \ref{fig:leakage_test_example} shows an example of how the metrics detect the data leakage.
Table \ref{tab:corruption_result} shows the results of leakage measurement. In addition to the two standard evaluation benchmarks (CEval~\citep{C_Eval} and MMLU~\citep{MMLU}), we conducted the same experiments on the alpaca dataset~\citep{alpaca} and the Alpaca-GPT4 dataset~\citep{alpaca_gpt4_instr}, which is likely used in the pre-training of large models, using its $\Delta L$ as reference. This demonstrates the unbiased nature and non-leakage of the OpsEval test set.

\subsection{Expert alignment of FAE-Score}
\label{sec:expert_alignment}

Table \ref{tab:gpt4_humaneval_corr} shows the correlation coefficients between various automated scoring metrics (ROUGE, BLEU, RAGAS, and FAE-Score) and Expert-Evaluation criteria. The results indicate that ROUGE and BLEU scores often misalign with Expert-Evaluation. This misalignment occurs because LLMs with poor performance may generate keywords that boost ROUGE and BLEU scores, while stronger LLMs might receive lower scores due to different wording from standard answers. While RAGAS \citep{es-etal-2024-ragas} aligns better with experts than ROUGE and BLEU, there is still a gap between its scoring rankings for different models and expert judgement standards. In contrast, FAE-Score rankings closely match Expert-Evaluation, particularly with the Accuracy metric. This suggests that FAE-Score is more reliable in assessing the factual accuracy of LLMs' outputs. Notably, GPT-4's performance in factual accuracy is reflected in its strong alignment with the Accuracy metric.
\section{Discussion}
\label{sec:discussion}

\subsection{Automated QA generation}

During the data collection process, we explored automating question-answer generation. Initially, we sampled QA pairs and manually evaluated their accuracy and domain relevance. Later, we utilized representative examples for few-shot learning, enabling GPT to generate and evaluate QA pairs automatically based on predefined criteria. 

\begin{figure}
    \centering
    \includegraphics[width=0.9\linewidth]{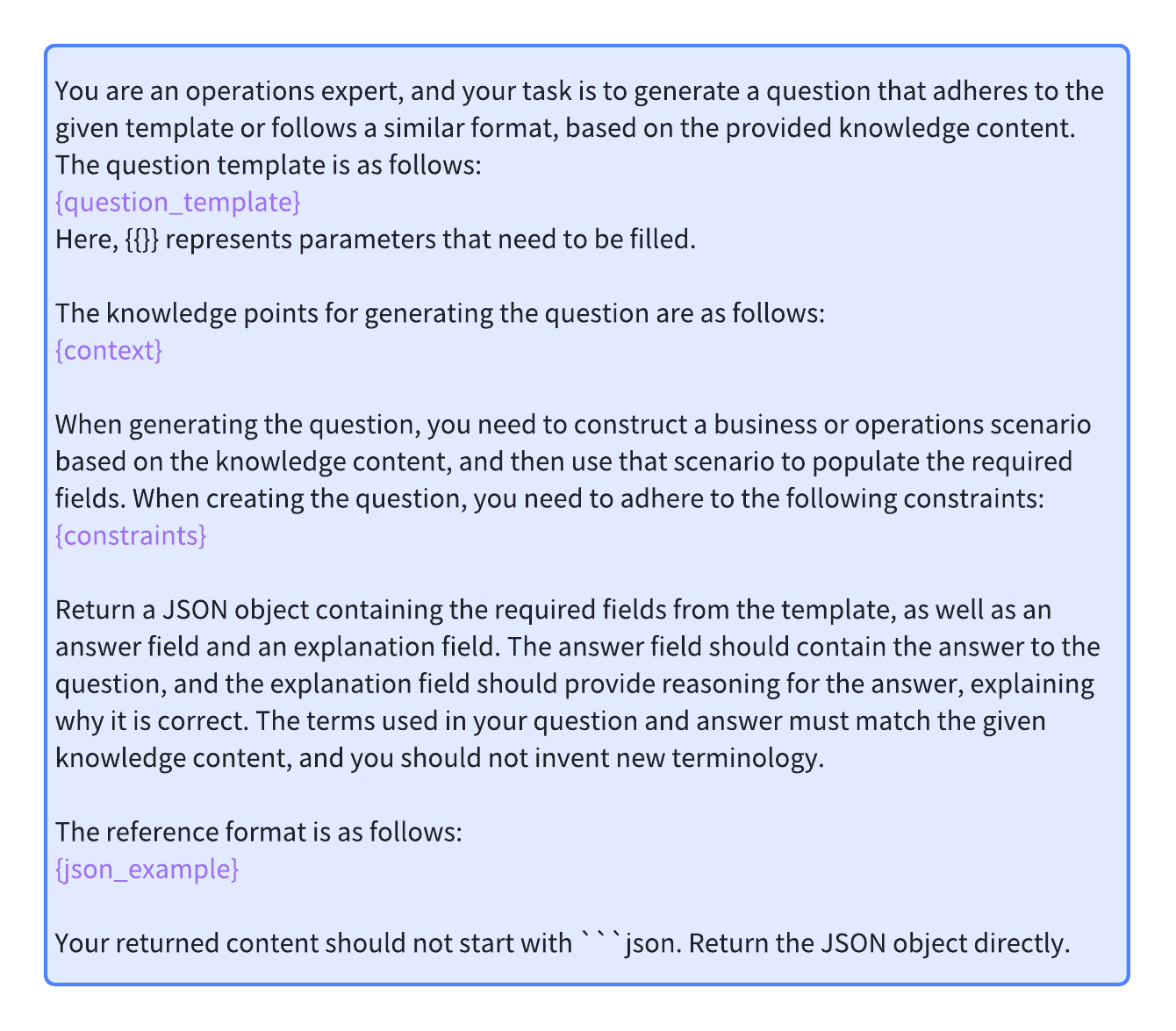}
    \caption{Prompt template for automated QA generation}
    \label{fig:qagen-prompt}
    \Description[Prompt template for automated QA generation]{Prompt template for automated QA generation}
\end{figure}

\begin{figure}
    \centering
    \includegraphics[width=0.9\linewidth]{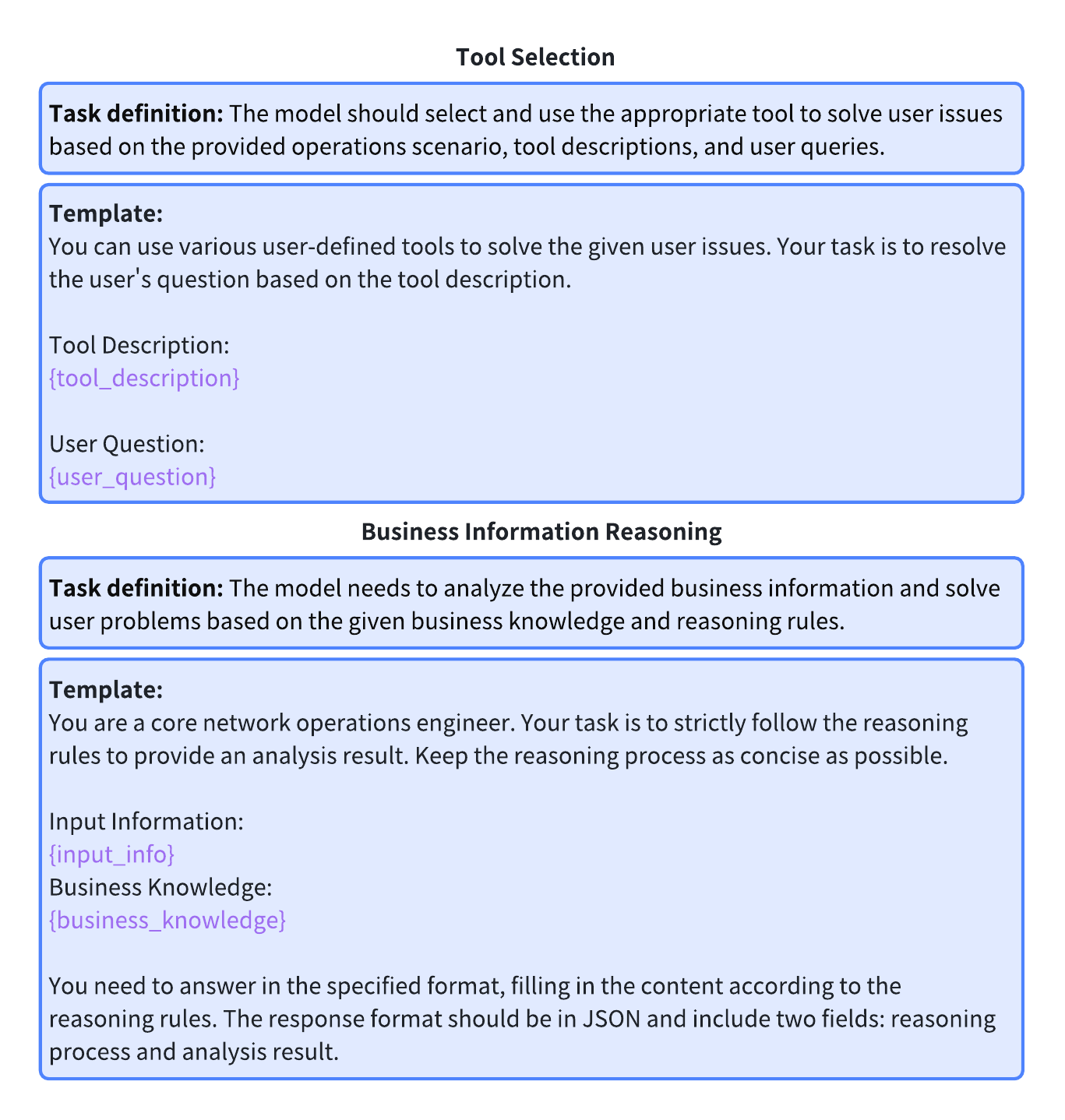}
    \caption{Some automatically generated QAs, their task description, template and example question}
    \label{fig:autogen-example}
    \Description[Some automatically generated QAs, their task description, template and example question]{Some automatically generated QAs, their task description, template and example question}
\end{figure}

Recognizing that most existing benchmarks focus primarily on simple knowledge-based questions, we designed various task-specific templates to address this limitation. These templates require the model to complete specific fields within the template using the provided knowledge content, rather than generating entire questions and answers. This prompt engineering approach allows us to generate detailed and context-specific Ops tasks based on extensive operational knowledge while improving the model’s instruction-following ability. By focusing on field-level completion, the overall structure of the QA remains consistent and accurate. 
Figure \ref{fig:qagen-prompt} shows the prompt template used for automatic QA generation, and Figure \ref{fig:autogen-example} illustrates some task cases.
This approach ensures a more diverse and comprehensive evaluation of model capabilities while maintaining the relevance and quality of generated tasks.

\subsection{Threats to Validity}
The internal validity of this study is primarily influenced by the deployment parameters and prompts for the LLMs. Variations in these configurations may impact the evaluation results, and while we strive to follow best practices, some optimizations may not fully reflect real-world settings.
The external validity is mainly limited by the datasets chosen. Our evaluation is based on specific datasets and Ops contexts, which may not generalize to other LLM deployment environments. In the future, we plan to expand OpsEval to include more datasets, scenarios, and deployment settings.

\subsection{Future Work}
\label{sec:limitation}

\textbf{Dataset Scale and Real-World Data.} While privacy constraints limit real-world company data, our ongoing collaborations aim to expand the dataset with practical scenarios. Expanding the dataset with real-world scenarios remains a key focus, while the benchmark prioritizes robust evaluation over dataset scale.
\textbf{Agent and RAG Introduction}: The inclusion of agents and Retrieval-Augmented Generation (RAG) techniques is constrained by the current large models' lack of foundational knowledge in operations. Our leaderboard will incorporate more complex tasks once open-source models possess sufficient operational capabilities.
\textbf{More Balanced Distribution.} While the current sub-domain distribution in our work attempts to reflect the varying importance of different topics within the industry, we are actively cooperating with more collaborators to achieve a more balanced distribution. 
\section{Related Works}
\label{sec:related-works}

\begin{table}[ht]
\caption{A comparison of OpsEval with other popular\\ datasets/benchmarks.}
\label{tab:related-works}
\footnotesize
\begin{tabular}{lccc}
        \toprule
        Dataset/Benchmark 
        & Ops Domain
        & Open-sourced 
        & Leaderboard \\
        \midrule
        MMLU & \xmark & \cmark & \cmark \\
        HELM & \xmark & \cmark & \cmark \\
        BIG-bench & \xmark & \cmark & \cmark \\
        SEAL & \ncmark & \xmark & \cmark \\
        CEval & \ncmark & \cmark & \cmark \\
        FLUE & \xmark & \cmark & \xmark \\
        MultiMedQA & \xmark & \cmark & \xmark \\
        CMB & \xmark & \cmark & \xmark \\
        NetOps & \ncmark & \xmark & \xmark \\
        OWL & \cmark & \xmark & \xmark \\
        OpsEval & \cmark & \cmark & \cmark \\
\bottomrule
\end{tabular}
\end{table}

\balance

As LLMs evolve rapidly, their complex and varied capabilities are increasingly recognized.
LLM specialized evaluation benchmarks can be divided into two categories: general ability benchmarks and domain-specific benchmarks.

\textbf{General ability benchmarks} assess the general abilities of LLMs across various tasks. These tasks evaluate LLMs' capacity for logical reasoning, general knowledge, common sense, and other similar abilities rather than being confined to a particular domain.
MMLU~\citep{MMLU} is a benchmark designed to measure knowledge acquired during pretraining by evaluating models exclusively in zero-shot and few-shot settings, covering 57 subjects across STEM.
HELM~\citep{HELM} employs seven distinct metrics in 42 unique scenarios, offering a comprehensive evaluation of LLMs' capabilities across multiple dimensions.
BIG-bench~\citep{BIG_bench} comprises 204 tasks spanning a wide array of topics, with a particular focus on tasks deemed beyond the reach of current LLMs.
SEAL~\citep{scale2024} features private, expert evaluations of leading frontiers models.
C-Eval~\citep{C_Eval} is a comprehensive Chinese evaluation suite designed to assess Chinese LLMs' advanced knowledge and reasoning abilities rigorously.

\textbf{Domain-specific benchmarks} evaluate the abilities of LLMs to handle tasks in specific fields. These benchmarks require LLMs to possess specialized knowledge in a specific domain and to respond in a manner consistent with the cognitive patterns of that field. Despite the rapid progression of LLMs in specialized domains, the evaluation metrics for these specific areas have received less attention.
FLUE~\citep{shah-etal-2022-flue} is an open-source comprehensive suite of benchmarks, including new benchmarks across 5 NLP tasks in financial domain.
MultiMedQA~\citep{MultiMedQA} is an extensive medical question-answering dataset, with questions derived from professional medical exams, research, and consultation records. 
CMB~\citep{CMB} includes multi-choice questions (CMB-Exam) and complex clinical questions based on real case studies (CMB-Clin).
NetOps~\citep{NetOps__DanLi} focuses on evaluations in the network field, which is relevant to the field of Ops. NetOps includes multi-choice questions in both English and Chinese and a few question-answering questions.
However, they only focus on wired network operations and while the dataset is released, they lack a benchmark that continuously updates the leaderboard.
OWL~\citep{guo2024owl} introduces Owl-Instruct and Owl-Bench datasets for IT operations, along with methods like HMCE for handling input length and a mixture-of-adapter for efficient tuning. However, it lacks a real-time updated leaderboard and does not provide a well-designed evaluation for IT operations QA tasks.

\section{Conclusion}
\label{sec:conclusion}
In this paper, we introduced \textbf{OpsEval}, the first comprehensive Ops benchmark suite designed for evaluating the performance of large language models (LLMs) in IT operations. We established a robust evaluation framework encompassing a wide range of sub-domains and tasks within Ops through rigorous data collection from multiple sources and meticulous preprocessing steps. Our benchmark includes a carefully selected set of \allquestioncount questions, which we have partially released to aid initial evaluations while protecting the integrity of the remaining dataset. It has undergone experiments in data leakage detection, ensuring its reliability.
Our observations, supported by quantitative and qualitative results, highlight the need for a balanced approach to selecting fundamental models, considering both performance and robustness. During the QA evaluation, the FAE-Score emerges as a more reliable metric than traditional metrics, suggesting its potential as a replacement for manual labeling in large-scale quantitative evaluations. Our failure rate analysis across 8 tasks and 3 abilities provides researchers with crucial insights and prospects for future breakthroughs.
The identified flexibility within the OpsEval framework presents opportunities for future exploration. This benchmark's adaptability facilitates the seamless integration of additional fine-grained tasks, providing a foundation for continued research and optimization of LLMs tailored for Ops. 
\begin{acks}




This work was partially funded by the National Key Research and Development Program of China (No.2022YFB2901800), the National Natural Science Foundation of China (62202445, 62272249, 62302244), the Beijing National Research Center for Information Science and Technology (BNRist) key projects, the Fundamental Research Funds for the Central Universities (XXX-63253249), and the National Natural Science Foundation of China-Research Grants Council (RGC) Joint Research Scheme (62321166652).

The research was conducted and the paper was written when the author Mingze Sun was undergraduate student in Tsinghua University and author Xidao Wen was postdoc in Tsinghua University.

\end{acks}

\newpage

\bibliographystyle{ACM-Reference-Format}
\bibliography{refs}
\balance



\end{document}